\documentclass{SCIS2025}
\newcommand{\eg}{\emph{e.g.}}
\newcommand{\ie}{\emph{i.e.}}
\newcommand{\name}{{\bf PILOT }}
\newcommand{\mame}{{\bf PILOT}}

\begin{document}
%\oa
%%%%%%%%%%%%%%%%%%%%%%%%%%%%%%%%%%%%%%%%%%%%%%%%%%%%%%%
%%% Authors do not modify the information below
%%% ×÷Õß²»ÐèÒªÐÞ¸Ä´Ë´¦ÐÅÏ¢
\ArticleType{NEWS \& VIEWS}
%\SpecialTopic{}
\Year{2025}
\Month{}
\Vol{68}
\No{}
\DOI{10.1007/s11432-024-4276-4}
\ArtNo{000000}
\ReceiveDate{9 April 2024}
\ReviseDate{12 July 2024}
\AcceptDate{17 January 2025}
\OnlineDate{}
\AuthorMark{Sun H-L}
\AuthorCitation{Sun H-L, Zhou D-W, Zhan D-C, et al}
%%%%%%%%%%%%%%%%%%%%%%%%%%%%%%%%%%%%%%%%%%%%%%%%%%%%%%%

%%% title: ±êÌâ
%%%   \title{title}{title for citation}
\title{PILOT: A Pre-Trained Model-Based \\Continual Learning Toolbox}{PILOT: A Pre-Trained Model-Based Continual Learning Toolbox}

%%% Corresponding author: Í¨ÐÅ×÷Õß
%%%   \author[number]{Full name}{{email@xxx.com}}
%%% General author: Ò»°ã×÷Õß
%%%   \author[number]{Full name}{}
\author[1,2]{Hai-Long Sun}{}
\author[1,2]{Da-Wei Zhou}{{zhoudw@lamda.nju.edu.cn}}
\author[1,2]{De-Chuan Zhan}{}
\author[1,2]{Han-Jia Ye}{{yehj@lamda.nju.edu.cn}}

%%% Authors' contribution. Í¬µÈ¹±Ï×
%\contributions{Equally contributed to this work.}

%%% Address. µØÖ·
%%%   \address[number]{Affiliation, City Postcode, Country}
\address[1]{School of Artificial Intelligence, Nanjing University, China}
\address[2]{National Key Laboratory for Novel Software Technology, Nanjing University, China}

\maketitle

%%%%%%%%%%%%%%%%%%%%%%%%%%%%%%%%%%%%%%%%%%%%%%%%%%%%%%%
%%% The main text. ÕýÎÄ²¿·Ö
%%%%%%%%%%%%%%%%%%%%%%%%%%%%%%%%%%%%%%%%%%%%%%%%%%%%%%%
\begin{multicols}{2}
\noindent The rapid advancements in deep learning have resulted in significant achievements across various fields. However, our ever-changing world often presents training data in a streaming format from an open environment. For example, while ChatGPT demonstrates exceptional inference capabilities, it struggles to provide users with the most up-to-date information. This challenge arises from the high costs associated with retraining a GPT model on new data daily. Therefore, the ability to continually update the model is critically important. Continual learning has been proposed as a solution to this challenge, allowing models to learn from streaming data. A major concern in continual learning is \emph{catastrophic forgetting}, where models forget previously learned information when acquiring new knowledge. Many methods have been developed to address this issue and enable models to learn from new data without forgetting former knowledge. In this paper, we focus on the Class-Incremental Learning (CIL) setting, which is a common scenario in continual learning.\looseness=-1

Traditional approaches assume that models are ``trained from scratch.'' However, with the rapid evolution of pre-training techniques, Pre-Trained Models (PTMs) have become widely used for downstream tasks. These PTMs are typically trained on extensive corpora or massive image datasets, resulting in robust generalizability. Consequently, research in CIL is shifting from training models from scratch to leveraging the power of PTMs.
According to a recent survey~\cite{zhou2024continual}, methods based on PTMs exhibit significantly superior performance compared to traditional methods relying on random initialization. This raises an important question: \textbf{Is there still a need to study traditional CIL?} To address this inquiry, we not only reproduce state-of-the-art methods in PTM-based CIL but also modify several traditional methods to be compatible with PTMs. This enables a fair comparison between PTM-based methods and traditional methods. \looseness=-1

We open-source the \textbf{P}re-tra\textbf{I}ned mode\textbf{L}-based c\textbf{O}ntinual learning \textbf{T}oolbox (\mame) for the machine learning community. It includes several traditional CIL approaches modified by PTMs and offers state-of-the-art algorithms to advance PTM-based CIL research. The source code of \name is available at \url{https://github.com/sun-hailong/LAMDA-PILOT}.

\lettersection{Compared with Other Toolkits} Since the current machine learning community lacks a toolbox that includes numerous PTM-based methods, there is an urgent need to develop a dedicated PTM-based toolbox. This toolbox will facilitate cutting-edge research and allow for fair comparisons between traditional methods and PTM-based methods using the same backbone. We primarily compare \name and other toolkits in the following three aspects:

{\noindent\bf Incorporation of PTMs.} \name not only encompasses traditional CIL methods but also extends support for the latest PTM-based CIL approaches. In contrast, other toolkits have mainly focused on conventional CIL methods and have not explored the integration of PTMs.

{\noindent\bf Network Architecture and Parameter Tuning.} By transitioning from the typical ResNet backbone to using PTMs, we design a unique parameter setting and tuning approach. While traditional toolkits can potentially be extended to accommodate PTMs, they are primarily designed with Convolution Neural Network (CNN). Hence, the parameters and hyper-parameter suited for CNN might not be optimal for PTMs.

{\noindent\bf Benchmarks and Datasets.} We provide benchmarks and datasets specifically curated for scenarios involving PTMs. These dedicated resources can play a pivotal role in obtaining accurate performance metrics and evaluations tailored to PTM-based CIL.

\begin{figure*}[t]
	
    \begin{center}
        \subfloat[CIFAR100, 10 Stages]
        {\includegraphics[width=0.46\columnwidth]{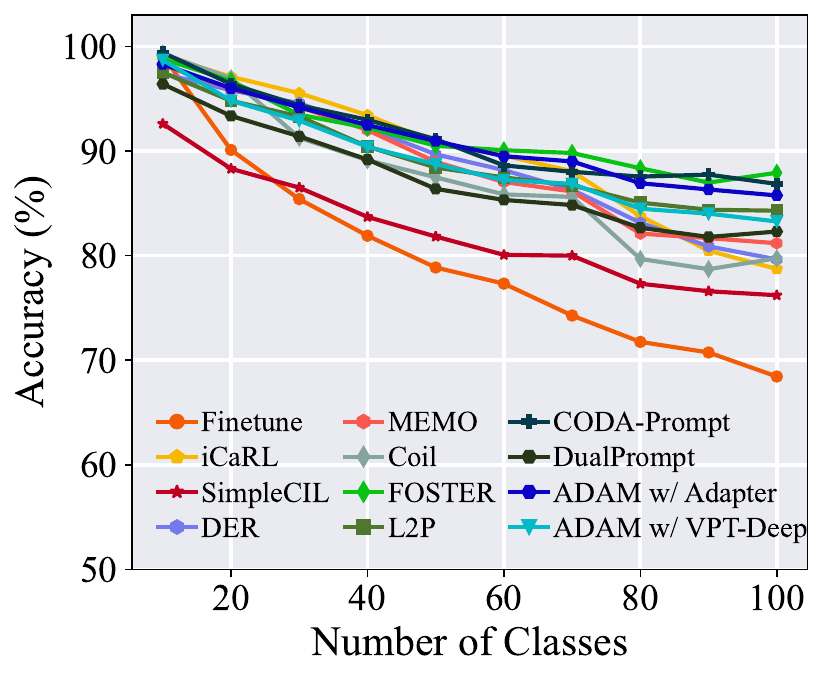}} 
        \subfloat[ImageNet-R, 10 Stages]
        {\includegraphics[width=0.45\columnwidth]{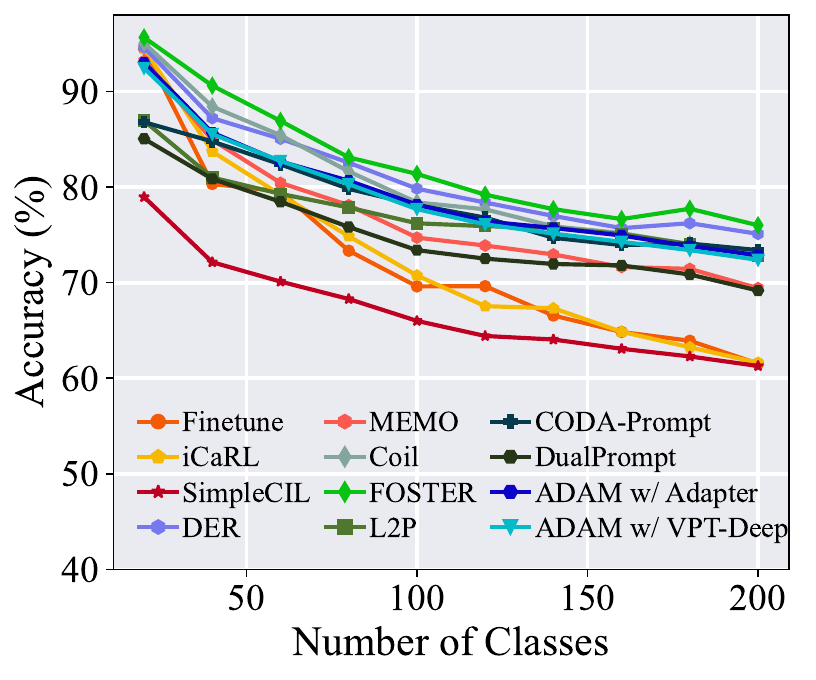}} 
        \subfloat[ImageNet-R, 10 Stages]
        {\includegraphics[width=0.45\columnwidth]{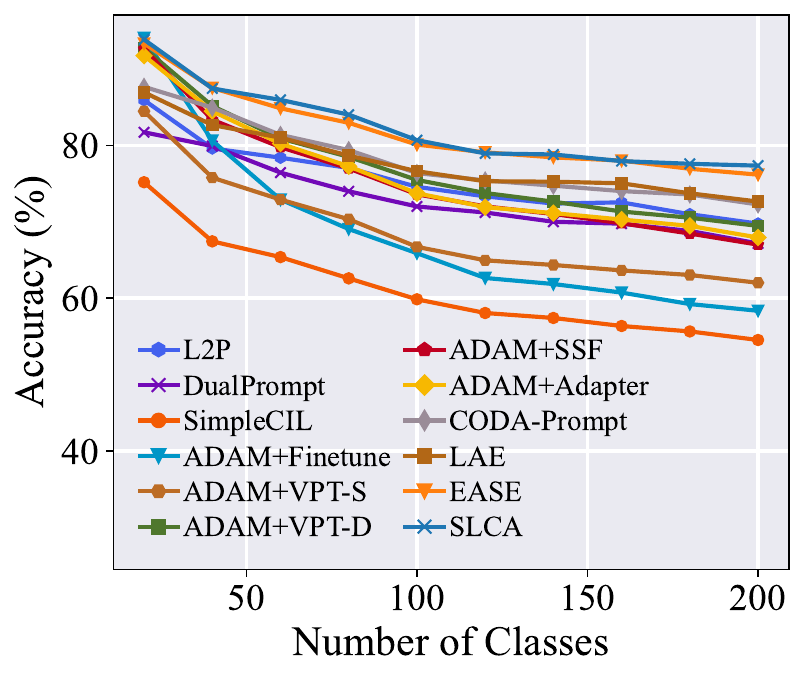}} 
        \subfloat[VTAB, 5 Stages]
        {\includegraphics[width=0.46\columnwidth]{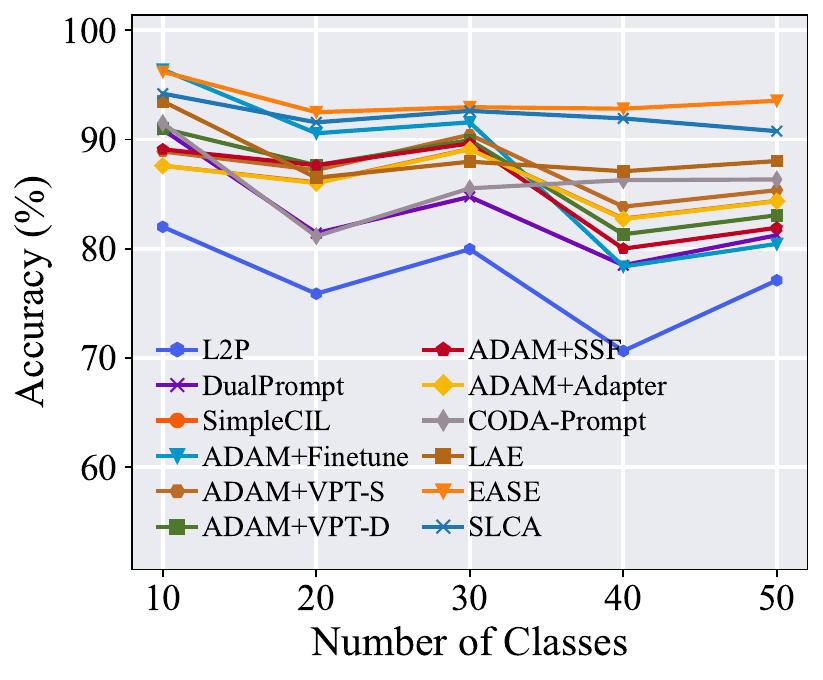}} 
    \end{center}
    
    \caption{  Reproduced incremental accuracy on CIFAR100, ImageNet-R, and VTAB. Subfigures (a) and (b) utilize the ViT-B/16-IN1K backbone, while subfigures (c) and (d) employ the ViT-B/16-IN21K backbone.} 
    \label{figure:results}
	
\end{figure*}

\lettersection{Implemented Algorithms}
In \mame, we implement 15 typical algorithms for CIL, including traditional methods modified by PTMs and PTM-based methods. Below, we list the latest PTM-based methods:
    {\bf Finetune} involves continually training a pre-trained model on new tasks. It updates all parameters and is vulnerable to severe catastrophic forgetting. 
    {\bf SimpleCIL}~\cite{zhou2024revisiting} constructs classifiers continually by extracting prototype features using PTMs, without the need for additional training on the downstream task.
    {\bf L2P}~\cite{wang2022learning} incorporates visual prompt tuning into CIL using a pre-trained vision transformer and establishes a prompt pool for selecting the instance-specific prompts.
    {\bf DualPrompt}~\cite{wang2022dualprompt} proposes two kinds of prompts based on L2P, \ie, general and expert prompts.
    {\bf CODA-Prompt}~\cite{smith2023coda} improves the prompt selection process with an attention mechanism.
    {\bf APER}~\cite{zhou2024revisiting}, based on SimpleCIL, employs parameter-efficient fine-tuning to acquire an adapted model. Subsequently, it concatenates the adapted model with the original one to obtain augmented features for constructing a prototype-based classifier.
    {\bf RanPAC}~\cite{mcdonnell2024ranpac} injects a frozen random projection layer with nonlinear activation to capture interactions between features with expanded dimensionality.
    {\bf SLCA}~\cite{zhang2023slca} improves the classification layer by modeling the class-wise distributions and aligning the classification layers in a post-hoc fashion.
    {\bf LAE}~\cite{gao2023unified} defines the online and offline learning protocol, where the online model is updated with cross-entropy loss, aiming to acquire new knowledge in new tasks. 
    {\bf EASE}~\cite{zhou2024expandable} designs an expandable subspace ensemble method for PTM-based CIL. \looseness=-1

\lettersection{Supported Datasets}: Due to the overlap in data between ImageNet-based benchmarks and the pre-trained dataset, ImageNet is not an appropriate choice for assessing PTM-based CIL methods, we provide some novel benchmarks for CIL which: 1) are entirely distinct from the ImageNet dataset, 2) present a significant domain gap from ImageNet, thereby challenging the PTM's ability to generalize, and 3) encompass large-scale datasets from various domains to establish a cross-domain class-incremental benchmark. On the other hand, since pre-trained models may possess extensive knowledge of upstream tasks, we evaluate performance on CIFAR100, CUB200, ImageNet-R, ImageNet-A, ObjectNet, OmniBenchmark, and VTAB. These datasets represent typical CIL benchmarks and include out-of-distribution datasets that exhibit a significant domain gap with ImageNet (\ie, the pre-trained dataset). Specifically, there are 50 classes in VTAB, 100 classes in CIFAR100, 200 classes in CUB, ImageNet-R, ImageNet-A, and ObjectNet, and 300 classes in OmniBenchmark. \looseness=-1

\lettersection{Evaluation Methodology} In CIL, a widely used performance metric is the test accuracy at each incremental stage, denoted as $\mathcal{A}_b$, where $b$ represents the stage index. Another important metric is the average accuracy across all stages, given by $\bar{\mathcal{A}}=\frac{1}{B} \sum_{b=1}^{B} \mathcal{A}_{b}$. In this work, we evaluate the incremental performance (Top-1 accuracy) at each stage, with results shown in Figure~\ref{figure:results}. 
We utilize datasets such as CIFAR100, ImageNet-R, ObjectNet, and VTAB, dividing all classes into several incremental stages. 
Due to some missing parameters in some papers (\eg, L2P), we have optimized a suitable parameter set for these methods. 
It is encouraging to observe that most re-implemented algorithms either match or exceed the performance benchmarks of the original publication. 
Moreover, we find that although traditional methods use PTM backbones and preserve some samples for replay, their performance is generally lower than PTM-based methods. This highlights the importance of leveraging pre-training techniques to design efficient CIL methods.

\lettersection{Conclusion}
We have introduced \mame, a pre-trained model-based continual learning toolbox. It includes a collection of reproduced PTM-based CIL methods and provides state-of-the-art algorithms for advanced research. \name aims to facilitate innovative research and development in the field of continual learning. In the future, we will continue to update our toolbox, expanding it to include more algorithms and datasets, and applying it to a wider range of settings.

%%%%%%%%%%%%%%%%%%%%%%%%%%%%%%%%%%%%%%%%%%%%%%%%%%%%%%%
%%% Acknowledgements. ÖÂÐ»
%%%%%%%%%%%%%%%%%%%%%%%%%%%%%%%%%%%%%%%%%%%%%%%%%%%%%%%
\Acknowledgements{This work is partially supported by  National Key R\&D Program of China (2022ZD0114805), NSFC (62476123, 62376118, 62006112, 62250069, 61921006), Fundamental Research Funds for the Central Universities (2024300373, 14380021), Key Program of Jiangsu Science Foundation (BK20243012), CCF-Tencent Rhino-Bird Open Research Fund RAGR20240101, the AI \& AI for Science Project of
Nanjing University, Collaborative Innovation Center of Novel Software Technology and Industrialization.}

%%%%%%%%%%%%%%%%%%%%%%%%%%%%%%%%%%%%%%%%%%%%%%%%%%%%%%%
%%% Supplements. ²¹³ä²ÄÁÏ, ·Ç±ØÑ¡
%%%%%%%%%%%%%%%%%%%%%%%%%%%%%%%%%%%%%%%%%%%%%%%%%%%%%%%
\Supplements{Appendix A.}

%%%%%%%%%%%%%%%%%%%%%%%%%%%%%%%%%%%%%%%%%%%%%%%%%%%%%%%
%%% Reference section. ²Î¿¼ÎÄÏ×
%%% citation in the content using "some words~\cite{1,2}".
%%% ~ is needed to make the reference number is on the same line with the word before it.
%%% Please make sure there are no more than 9 items of references.
%%%%%%%%%%%%%%%%%%%%%%%%%%%%%%%%%%%%%%%%%%%%%%%%%%%%%%%
\bibliography{scis}
\bibliographystyle{unsrt}

\end{multicols}
\end{document}